\documentclass[letterpaper, 10 pt, conference]{ieeeconf}  

\IEEEoverridecommandlockouts                              

\usepackage[pdftex]{graphicx}
\usepackage{graphics} 
\usepackage{epsfig} 
\usepackage{times} 
\usepackage{amsmath} 
\usepackage{amssymb}  
\usepackage{xcolor}
\usepackage[ruled,vlined]{algorithm2e}
\usepackage{float}
\usepackage{hyperref}
\usepackage{multirow}
\usepackage{subcaption}

\usepackage{pgffor}
\usepackage{tabularx}

\newcolumntype{Y}{>{\centering\arraybackslash}X}

\definecolor{burntorange}{rgb}{0.8, 0.33, 0.0}
\definecolor{oceanboatblue}{rgb}{0.0, 0.47, 0.75}

\newcommand{\etal}{\textit{et al.}}

\newcommand {\rfig}[1]{Figure~\ref{#1}}

\title{\LARGE \bf
ZePHyR: Zero-shot Pose Hypothesis Rating
}

\author{Brian Okorn*, Qiao Gu*, Martial Hebert and David Held
\thanks{* indicates equal contribution.}
\thanks{B. Okorn, Q. Gu, M. Hebert and D. Held are with the Robotics Institute, Carnegie Mellon University, 5000 Forbes Ave, Pittsburgh, PA 15213. ({\tt\small bokorn@andrew.cmu.edu, qiaog@andrew.cmu.edu, hebert@cs.cmu.edu, dheld@andrew.cmu.edu})}
}

\begin{document}

\maketitle

\thispagestyle{empty}
\pagestyle{empty}


\begin{abstract}
    Pose estimation is a basic module in many robot manipulation pipelines. Estimating the pose of  objects in the environment can be useful for grasping, motion planning, or manipulation.  However, current state-of-the-art methods for pose estimation either rely on large annotated training sets or simulated data. Further, the long training times for these methods prohibit quick interaction with novel objects.  To address these issues, we introduce a novel method for zero-shot object pose estimation in clutter.  Our approach uses a hypothesis generation and scoring framework, with a focus on learning a scoring function that generalizes to objects not used for training.  We achieve zero-shot generalization by rating hypotheses as a function of unordered point differences.  We evaluate our method on challenging datasets with both textured and untextured objects in cluttered scenes and demonstrate that our method significantly outperforms previous methods on this task. We also demonstrate how our system can be used by quickly scanning and building a model of a novel object, which can immediately be used by our method for pose estimation. Our work allows users to estimate the pose of novel objects without requiring any retraining. Additional information can be found on our website \href{https://bokorn.github.io/zephyr/}{https://bokorn.github.io/zephyr/}
\end{abstract}



\section{Introduction} \label{sec:intro}



6D pose describes the position and orientation of an object, defined in a reference frame relative to a predefined model of the object. An object's 6D pose fully describes the state of a static rigid object and, as such, is commonly used as a representation for planning~\cite{dantam2016incremental,zucker2013chomp}. A robot can use an estimate of an object's pose to perform complex manipulation interactions with the object~\cite{kim2016planning,thomas2018learning,goldfeder2009columbia, ciocarlie2014towards}.

Current state-of-the-art methods for object pose estimation train a new model for each object they are being evaluated on~\cite{xiang2017posecnn, wang2019densefusion, Brachmann2014learning}. This requires a large amount of annotated training data, either produced by capturing and annotating large datasets or through rendering the object in synthetically generated scenes. For example, the YCB-Video dataset~\cite{xiang2017posecnn} contains 133,827 human-annotated images with roughly 25,000 images per object.  Although this dataset has enabled the training of powerful deep learning methods~\cite{xiang2017posecnn, wang2019densefusion}, curating such a human-labeled dataset (including both capturing a diverse dataset and labeling the data) for each new object that a robot must interact with is cumbersome. Methods that rely on purely simulated data~\cite{tremblay2018dope,deng2019poserbpf,sundermeyer2020augmented} avoid this limitation but must instead contend with the sim2real gap between the synthetic data and real sensor observations. Improved rendering~\cite{tremblay2018falling} and domain randomization techniques~\cite{dwibedi2017cut} have been suggested to alleviate this gap, but ensuring that the simulated data accurately represents the variations observed in the real world continues to be an open problem. 

\begin{figure}
    \centering
    \includegraphics[trim={180 20 208 140}, clip,width=0.49\columnwidth]{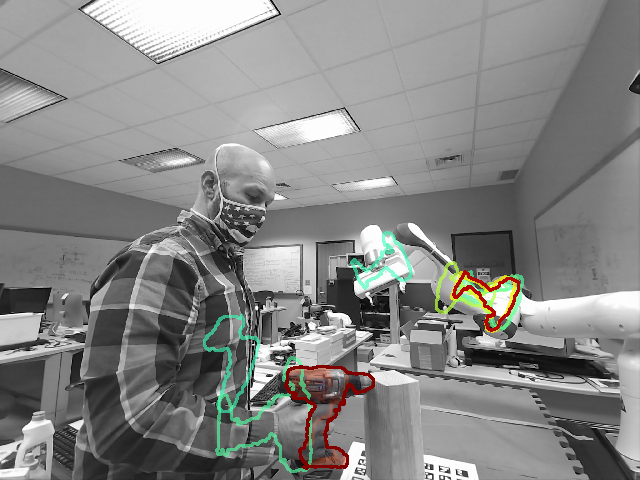}
    \includegraphics[trim={290 120 130 80}, clip,width=0.49\columnwidth]{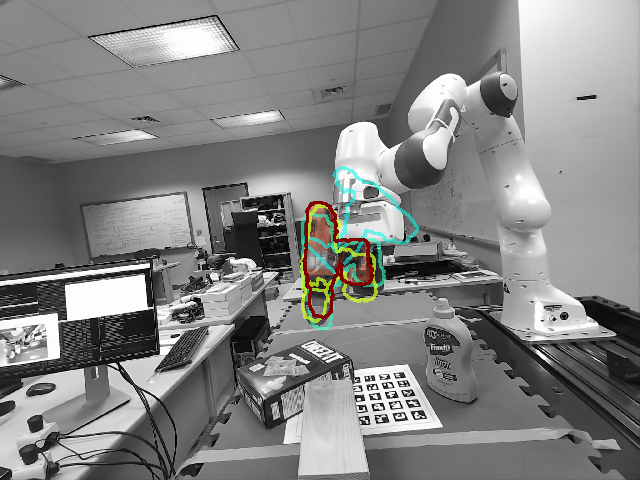}
    \caption{Pose hypotheses scored using Zero-shot Pose Hypothesis Rating on novel drill object, reconstructed at test time. The highest scoring pose is rendered in color. Poses are outlined in color corresponding to score, with highly-rated poses in \textcolor{red}{red} and to lower ones in \textcolor{cyan}{blue}.
    }
    \label{fig:Figure1}
    \vspace{-2em}
\end{figure}

Regardless of how this data is obtained, training new networks has a time and space cost. This training can take many hours, which prevents robots  using such systems from quickly being able to interact with new objects. Additionally, new network weights are trained for each new object, which presents a difficulty for memory-constrained robot systems. These constraints do not scale well in cases where robots need to interact with many different types of objects.

One approach to mitigate these issues is to use a non-learned geometry-based method~\cite{drost2010ppf, vidal2018method}.  
These methods, however, do not typically capture visual texture well, 
and they rely on hard-coded, rather than learned, invariances, which limits the potential accuracy of the system (based on our experiments in Section \ref{sec:results-main}).
A few recent learning-based approaches have attempted to perform zero-shot object pose estimation~\cite{park2019latentfusion,sundermeyer2020multipath} but these methods require instance segmentation masks to be provided as input, which limits their use in a ``zero-shot" system, as such masks are typically trained per-object.




We seek to remove these limitations by developing a novel learning-based method for zero-shot object pose estimation that can handle both textured and untextured objects in cluttered scenes and does not require object masks as input. Our method uses the paradigm of pose hypothesis generation and evaluation: given a scene, a large number of candidate poses consistent with the observation are generated. The fitness of each hypothesis is then evaluated and the best-fit candidate is selected. Such an approach requires the hypothesis rating function to give appropriate weight to the features that most correlate with the correct pose. The variation between sensor data and the object model, caused by sensor noise or lighting changes, as well as partial occlusions, can make designing this scoring function challenging. Past approaches to hypothesis scoring have used voting over hypotheses or feature matching~\cite{fischler1981random,brachmann2017dsac,drost2010ppf}; 
in contrast, this paper proposes a scoring function that learns to compare the observed images and rendered model points.  Our learned scoring function demonstrates a significant improvement on zero-shot object pose estimation over a wide set of objects and environmental variations. 




The key insight of our method is to use a learned scoring function that compares the sensor observation to a sparse rendering of each candidate pose hypothesis. This scoring function receives as input \emph{an unordered set of point differences}, shown in~Fig.~\ref{fig:pipeline}, which we show is crucial to perform zero-shot generalization to novel objects not seen in the training set.  Our method is trained over a disparate set of objects and then evaluated on novel objects not included in the training set. 
 

We demonstrate that our Zero-shot Pose Hypothesis Rating method (ZePHyR) works on objects in clutter without requiring object masks as input, unlike past zero-shot methods~\cite{sundermeyer2020multipath,park2019latentfusion}.  ZePHyR handles both untextured objects as well as objects with significant visual texture, not seen at training time. Therefore, ZePHyR achieves the goal of zero-shot object pose estimation mentioned earlier: 
\begin{itemize}
\item We require no new human annotations or large-scale synthetic data generation to interact with novel objects.
\item We require no retraining for novel objects.  
\item ZePHyR uses only a single set of network weights, rather than requiring new weights for each unique object, reducing the memory constraints. 
\end{itemize}
We evaluate our method on YCB-Video and LineMOD-Occlusion, two challenging pose estimation datasets. Our method achieves state-of-the-art results over previous zero-shot pose estimation methods. 



\vspace{-0.4em}
\section{Related Work}
\vspace{-0.1em}
\subsection{Non-learned Zero-shot Pose Estimation}


Zero-shot pose estimation is the task of estimating the pose of objects not seen at training time. Non-learning based approaches \cite{besl1992icp, Ulrich2012combining, Hinterstoisser2012gradientresponse, Hinterstoisser2012modelbased, munoz2016fast6d, Joseph2014fpm, munoz2016fast6dforests, guo2019fastglobal, yu2019robust} are inherently zero-shot, leveraging robust features and the available object model at test time. 
Point Pair Features (PPF)~\cite{vidal2018method, drost2010ppf, drost20123d,kim2011object,drost20123d,hinterstoisser2016going} use pairs of oriented points to generate geometrically consistent pose hypotheses and select the best hypothesis using voting and clustering. These are the top-performing zero-shot methods on the BOP leader board~\cite{hodan2018bop}, when averaged over all datasets, but struggle to compete with deep learned methods on the highly textured YCB dataset due to the methods being exclusively based on depth. 

\subsection{Learned Zero-shot Object Pose Estimation}

Several learned methods solve the zero-shot pose estimation problem using class-based pose estimation~\cite{manuelli2019kpam, wang2019normalized} as opposed to instance-based pose estimation. These methods learn a pose estimator capable of generalizing among objects in the given class, but such methods are not intended to generalize to novel classes. While this is a step in the direction of zero-shot pose estimation, it still requires training a new network for each class. 

Pose refinement methods like DeepIM~\cite{li2018deepim} learn to estimate the residual pose between the observed data and a rendered viewpoint and have shown to generalize well to unseen classes of objects. These methods, however, require the initial rendered pose to be relatively close to the observation to produce accurate results, and as such is primarily used to refine a coarse pose prediction. Our method requires no such close initialization.

A few recent zero-shot methods use a learned representation of the object in their pose estimation pipeline~\cite{xiao2019pose,sundermeyer2020multipath,park2019latentfusion}. While these methods have been shown to generalize across objects, they require a bounding box for the target object, which is obtained using an object-specific learned detector (and hence not a zero-shot system) or the ground-truth bounding box. 
This requirement is avoided in the MOPED dataset~\cite{park2019latentfusion}, as there is only a single object in the scene, which greatly simplifies the task of estimating the object mask~\cite{xie2020best}. For the LineMOD-Occlusion dataset, ground truth object masks are used~\cite{park2019latentfusion}. Our method does not require such bounding boxes or masks as input, making it truly zero-shot.

\begin{figure*}[!ht]
    \centering
    \includegraphics[trim={0 30 0 5}, clip, width=\textwidth]{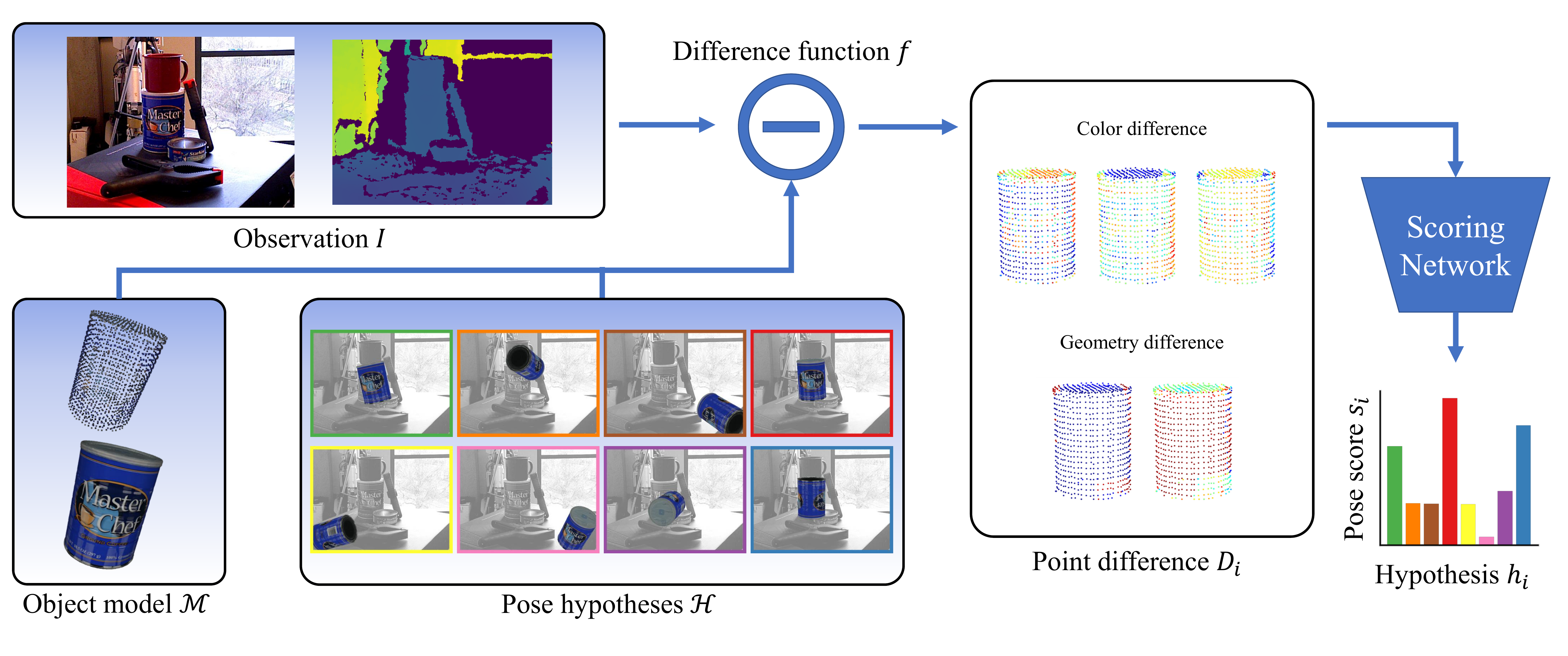}
    \caption{System Pipeline. Our method first projects the sampled model points $\mathcal{M}$ onto the observation $I$ according to a pose hypothesis $h_i$. Then $D_i$ are extracted as the point-wise differences between the observation and the projected model points, describing the alignment of the pose hypothesis at each projected point. A network takes in $D_i$ and regresses to a score $s_i$ for each pose $h_i$ which evaluates how well the pose matches the observation. }
    \label{fig:pipeline}
 	\vspace{-2em}
\end{figure*}

\subsection{Pose Scoring}

There has been some study of learned fitness functions. Differentiable RANSAC (DSAC)~\cite{brachmann2017dsac} explores learning a fully differentiable RANSAC algorithm. Specifically, they study the use of a REINFORCE style loss for scoring candidate hypotheses. We take inspiration from this work; however, their method focuses on a different task of camera localization rather than object pose estimation; as a result, many important details of our method, such as the input featurization and network architecture, are significantly different from their approach. Pose Proposal Critic~\cite{brynte2020pose} learns to regress to the reprojection error between a rendered pose and the observation. They numerically differentiate this error function as a means of pose refinement. However, they only evaluate this approach as a pose refinement technique, with a close initial pose estimate; in contrast, our focus is on evaluating a large set of pose hypotheses that span the entire observation space.

\vspace{-0.5em}
\section{Method}
\vspace{-0.5em}

\subsection{Overview} \label{sec:method-overview}

The primary objective of this work is zero-shot object pose estimation in clutter. To achieve this, we train our pose estimation method on one set of objects and then evaluate on a set of novel objects, without requiring any re-training. This differentiates our work from previous work that requires real or synthetic training data of the test objects~\cite{xiang2017posecnn, wang2019densefusion}.  Our work additionally differentiates from other zero-shot pose estimation work~\cite{park2019latentfusion, sundermeyer2020multipath,vidal2018method}, in that it operates well in cluttered scenes, requires no object masks as input, and produces accurate poses for both textured and untextured objects. 

An overview of our method is shown in \rfig{fig:pipeline}. Given a set of 6D pose hypotheses, we first project each hypothesis into the scene.  Our method learns to score each hypothesis by comparing differences in the projected object model point cloud to the RGB-D observation. 
For each projected model point, we extract the color and geometry information from both the model and the observation and compute the local differences of the extracted information. This yields a set of \emph{point-differences}, one for each projected model point. Each element in this set encodes the local alignment between the model and the observation with respect to color and geometry. We adopt a point-based  network~\cite{Qi2017pointnet, qi2017pointnet2} to analyze this unordered set of point-differences and regress to an overall score for each pose hypothesis. Focusing on differences as well as adopting a point-based neighborhood structure helps us avoid overfitting to object-specific properties from the training set and allows us to generalize to unseen objects at test time. 





In this work, our primary focus is the learned scoring function and we use existing methods to generate our initial pose hypothesis set. While many algorithms could be used to generate these potential object pose hypotheses~\cite{brachmann2017dsac,narayanan2016perch,collet2011moped}, we use a combination of Point Pair Features~\cite{drost2010ppf} and SIFT features~\cite{lowe1999object}.



\subsection{Learned Scoring Function}




The main goal of our method is to score pose hypotheses by projecting them into the observed scene and learning to compare their local geometric and color differences.
Suppose that we have a set of 6D pose hypotheses $\mathcal{H}=\{h_i\}_{i=1}^m$ that we wish to evaluate. 
We represent the object as a point cloud $\mathcal{M}=\{x_j\}_{j=1}^n$, sampled from the provided object mesh model, or obtained from a 3D reconstruction pipeline. Each point contains both geometric (depth and normal)  and color information drawn from its local region on the object. Similarly the observation image $I$ contains geometric and color values from the observation. To evaluate hypothesis $h_i$, we project each object point $x_j$ onto the observation's image plane, using the known camera parameters. This projection gives a point at image coordinates $y_{ij}$ with transformed point values  $\tilde{x}_{ij}$ (the point depth and normal vector are transformed; the color of the projected point is unchanged).
For each pose hypothesis, the difference between the projected values, $\tilde{x}_{ij}$, and their corresponding image values, $I(y_{ij})$, is computed according to a simple distance function, $d_{ij} = f(\tilde{x}_{ij}, I(y_{ij}))$ (see Section~\ref{sec:feats} for details). 

The set  $D_i=\{d_{ij}\}_{j=1}^m$ represents an unordered set of point differences for pose hypothesis $h_i$, each of which is associated with a given point $x_j$ in the model and a location $y_{ij}$ in the observation image.
We train a deep neural network $g_\theta(D_i)$ with parameters $\theta$ to analyze this difference set and regress to a pose fitness score, $s_i$. While one might assume that a simple hand-designed function for $g$ would be sufficient, in practice, however, occlusions, lighting differences and other confounding factors make such simple methods ineffective. 
Our learned function can intelligently combine point differences on multiple parts of the object to robustly estimate the most likely pose hypothesis.


\subsection{Loss Function}

To train this hypothesis scoring function, we adopt the probabilistic selection loss proposed by DSAC~\cite{brachmann2017dsac}, as it directly optimizes the expected pose error when hypotheses are sampled according to the predicted scores. 
For each pose hypotheses $h_i$ with corresponding true pose error $\epsilon_i$, we compute the expected pose error of sampling according to the softmax distribution induced by $s_i$, ${\mathcal{L}=\sum_{i=1}^m softmax(s_i) \epsilon_i}$.


In our experiment, $\epsilon_i$ is defined as the log of the average point distance (ADD) for non-symmetric objects and its symmetric analog (ADD-S) for symmetric ones \cite{xiang2017posecnn}. Empirically, we find that the log of this error better dampens the effects of outliers. More discussion can be found in Section~\ref{sec:feats}. At test time, the highest-scoring pose hypothesis is selected. The inference pipeline is described in Algorithm \ref{alg:pipeline}.
\vspace{-1em}
\begin{algorithm}
\caption{Hypothesis Scoring Pose Estimation}
\label{alg:pipeline}
\SetAlgoLined
 Compute initial pose hypothesis set $\mathcal{H}=\{h_i\}_{i=1}^m$\;
 \ForEach{ $h_i$ in $\mathcal{H}$}{
 Project all model points according to $h_i$ onto the image plane to get projected model points  $\tilde{x}_{ij}$ at projected image coordinates $y_{ij}$\;
 Get observation points $I(y_{ij})$\;
 Compute point differences $d_i = f(\tilde{x}_{ij}, I(y_{ij}))$\;
 Score point-differences $s_i=g_\theta(\{d_{ij}\}_{j=1}^m)$\;
 }
Return hypothesis $h_{i^*}$, where $i^* = \arg\max_{i} s_i$\;
\end{algorithm}
\vspace{-1.0em}

\subsection{Implementation details}
\subsubsection{Hypothesis Generation}
We generate the initial hypotheses set using the commercially available Point Pair Feature software, HALCON 20.05 Progress software~\cite{halcon}, which implements the PPF algorithm described in Drost~\etal~\cite{drost2010ppf}. For each observation, we use the top 100 pose hypotheses generated by PPF. For detecting objects with high visual texture (e.g. for all objects in YCB-V), we augment these hypotheses using Dense SIFT feature matching. We obtain pose hypotheses from these features by aligning the surface normals and SIFT orientations of pairs of matched SIFT features; aligning the SIFT orientations and normals enables a single pair of matched SIFT features to define a 6D pose hypothesis. 

\subsubsection{Network Input}
\label{sec:feats}
As input to the hypothesis scoring function, we use very simple geometric and color information for both the model and observation data. For each point on the model, we compute its 3D location, surface normal, and color in HSV space. When projecting each point into the observation frame, we transform both the normals and 3D coordinates to compute the depth and normal with respect to the camera. The color data is unaffected by the projection. Similarly, we compute local surface normals from the observation, and thus we obtain depth, normal and HSV color information at each pixel of the observation image. 

To create the network inputs $d_{ij}$, we compute the signed difference between the projected and observed points for both depth and color. For surface normals, we use the cosine of the angle difference between the projected and observed normals. Additionally, we concatenate the projected image coordinates, $y_{ij}$, of the associated image point, normalized to zero mean and unit variance, as an additional input, to provide the structural neighborhood information. 

\subsubsection{Network Structure}

Our network takes in the set of point-differences $D_i=\{d_{ij}\}_{j=1}^m$  and outputs a single score, $s_i$, that estimates how well the pose hypothesis $h_i$ matches the observation. Because $D_i$ is an unordered set of point-differences, we use a network architecture designed to handle unordered sets of points;  specifically, we use PointNet++~\cite{qi2017pointnet2}. Our experiments show that the loose neighborhood structure of this architecture enables zero-shot generalization to unseen objects.
To define the spatial neighborhood for grouping points in PointNet++'s point set abstraction layers, we use the normalized image coordinates. We explore the effect of networks with different neighborhood structures in Section~\ref{sec:neighborhood}. 
See the supplementary material for more experimental details and hyperparameters.

\begin{table*}[ht!]
\small
\centering
\begin{tabular}{|c||c|c|c|c||c|c|}
\hline
 & \multicolumn{4}{c||}{Zero-Shot Methods} & \multicolumn{2}{c|
 }{Object Specific Methods} \\
 & \multirow{2}*{Drost~\cite{drost2010ppf}} & \multirow{2}*{Vidal~\cite{vidal2018method}} & \multirow{2}*{Multipath~\cite{sundermeyer2020multipath}} & ZePHyR + & \multirow{2}*{CosyPose~\cite{labbe2020}} & \multirow{2}*{Pix2Pose~\cite{Park_2019_ICCV}} \\
 &  & & & Drost (Ours) & & \\
\hline
YCB-V &  0.344 & 0.450 & 0.289 & \textbf{0.516} & 0.861	& 0.675 \\
LM-O &  0.527 & 0.581 & 0.217 & \textbf{0.598} & 0.714 & 0.588 \\
\hline
\end{tabular}
\caption{AR scores for methods of zero-shot and object specific pose estimation on object pose datasets (higher is better). }
\label{tbl:main_results}
\vspace{-2em}
\end{table*}

\section{Experiments}

\subsection{Datasets}
\label{sec:Datasets}

We evaluated our method on two of the most popular datasets in the BOP Challenge~\cite{hodan2018bop}, the YCB-Video (YCB-V) dataset~\cite{xiang2017posecnn} and the LineMOD-Occlusion (LM-O) dataset~\cite{Brachmann2014learning}. In these experiments, we follow the evaluation protocol set up by the BOP Challenge, with the additional constraint that our method is not trained on the objects it is tested on. This allows us to test our ability to perform zero-shot generalization to novel objects.

\textbf{YCB-Video dataset} (YCB-V)~\cite{xiang2017posecnn} contains 92 RGB-D video sequences of 21 YCB objects \cite{calli2015ycb} of varying shape and texture, annotated with 6D poses. This a particularly challenging dataset for object pose estimation due to its varying lighting conditions, occlusions, and sensor noise. We follow the dataset split in \cite{xiang2017posecnn}, and for the evaluation, we adopt the BOP testing set \cite{hodan2018bop}, where 75 images with higher-quality ground-truth poses from each of 12 test videos are used. 
To demonstrate the generalization ability of our method, one half of the objects are used for training, and the other half are used for testing. To accommodate the full dataset, a second network is trained with train and test objects exchanged, such that each network only sees half of the objects during training, and no network is trained on the objects that it will be tested on. 
Note that we train our network on the training (seen) objects in the YCB-V training split and test on the testing (unseen) objects in the testing split, so not a single test image or object is seen during training. 
When evaluating on YCB-V, we use hypotheses generated form both PPF and SIFT matching to handle the high degree of visual texture. We also adopt a ICP refinement step~\cite{besl1992icp} for post-processing.

\textbf{LineMOD-Occlusion dataset} (LM-O) \cite{Brachmann2014learning} adopted a single scene from the test set of the larger LineMOD (LM) dataset \cite{Hinterstoisser2012modelbased} and provides ground-truth 6D pose annotations for 8 low-textured objects. For training, we used the PBR-BlenderProc4BOP \cite{hodan2019photorealistic} training images provided by the BOP challenge. This dataset contains photo-realistic synthetic images of LM objects dropped onto a table, with randomized background texture and object materials. Our model is only trained on synthetic images of the 7 objects that are in the LM dataset but \textit{not} in the LM-O dataset; we then evaluate on the LM-O objects, which were not seen at training time. When evaluating on LM-O, we only use hypotheses generated by PPF; we find that SIFT hypotheses are ineffective on this dataset since the objects do not contain much visual texture.

\subsection{Metrics}
\label{sec:Metrics}
As suggested by the BOP challenge, we report the average recall (AR) scores as the average of the following three average-recall pose error metrics: Visible Surface Discrepancy (VSD), Maximum Symmetry-Aware Surface Distance (MSSD), and Maximum Symmetry-Aware Projection Distance (MSPD). For a detailed formulation of each metric, please refer to the supplementary material and~\cite{hodan2018bop}. 

\subsection{Baselines}

We compare our method to both zero-shot and object-specific methods. As we are most concerned with our performance as compared to other zero-shot methods, we compare to two variants of Point Pair Features, Drost~\cite{drost2010ppf} and Vidal~\cite{vidal2018method}. An implementation of Drost's PPF~\cite{halcon} is used as the hypothesis generation algorithm in our work. Vidal had until recently been the top-performing method in the BOP challenge, and demonstrates the peak performance of PPF-only systems (although their code is not available). 
Other recent papers have proposed learning-based methods for zero-shot pose estimation, namely Multipath Augmented Autoencoders~\cite{sundermeyer2020multipath}, which we compare against. While this method has been shown to generalize to unseen objects, the reported results that we include are a product of training a single model on the test objects; further, their method utilizes an object-specific detection network (also trained on the test objects)~\cite{he2017mask}. 
In addition to the zero-shot baselines, we report the current state of the art in object-specific methods as CosyPose~\cite{labbe2020} and Pix2Pose~\cite{Park_2019_ICCV}. Both of these methods train a network on annotated instances of the test objects and have weights specifically associated with each object. While we are not attempting to match the performance of these systems, we report their results to illustrate the still remaining gap between zero-shot and object-specific methods. 

\subsection{Zero-shot Pose Estimation Results}
\label{sec:results-main}
\begin{figure*}[ht!]
    \captionsetup[subfloat]{labelformat=empty}
    \centering
    \subfloat[Input Image]{
    \includegraphics[trim={0 0 0 0}, clip,width=0.27\textwidth]{figures/000051_001699_img.png}
    }
    \hfill
    \subfloat[Original PPF Results]{
    \includegraphics[trim={0 0 0 0}, clip,width=0.27\textwidth]{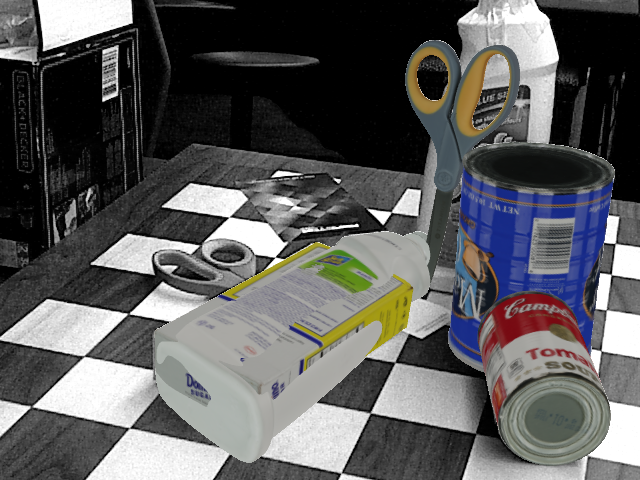}
    }
    \hfill
    \subfloat[Our Improved Results]{
    \includegraphics[trim={0 0 0 0}, clip,width=0.27\textwidth]{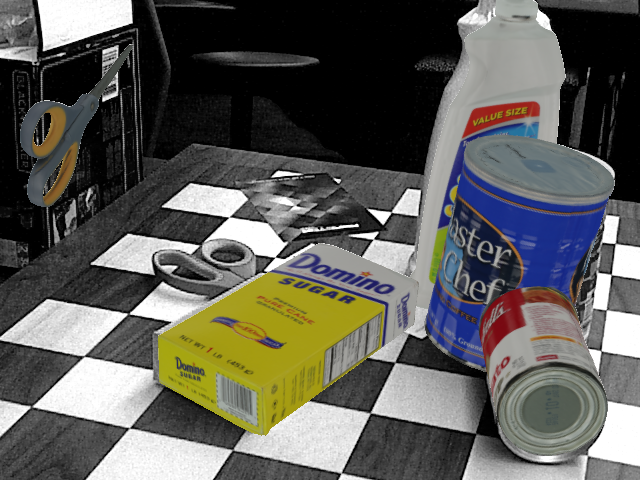}
    }
    \vspace{-0.2em}
    \caption{Qualitative results on image from YCB-V dataset showing the improved accuracy of our method.}
    \label{fig:qual_results}
    \vspace{-1.2em}
\end{figure*}

In Table~\ref{tbl:main_results}, we find that our method outperforms all zero-shot methods, significantly improving over our initial pose hypotheses produced by Drost and outperforming the best PPF-only solution in Vidal~\cite{vidal2018method}. We see the largest improvement on the YCB dataset, where PPF is unable to fully resolve the pose of the geometrically symmetric but textually asymmetric objects, seen in failure to match the cylindrical objects in Figure~\ref{fig:qual_results}. Our method is able to leverage both color and geometry, selecting the most accurate pose hypothesis.  Additionally, we find our method to be comparable to the object-specific results produced by Pix2Pose~\cite{Park_2019_ICCV}.
While DeepIM~\cite{li2018deepim} is a local refinement method, and not directly comparable to ZePHyR, we do evaluate its performance based on PPF in the supplementary material.


\subsection{Evaluating Generalization}

As we stated previously, in order to ensure our network is not trained on the test objects we split the objects in YCB-V into two halves, training a network on each set of objects. We select via index parity, as it separates the dataset into splits with roughly equal numbers of symmetric and asymmetric objects, with ``Object Set 1'' and ``Object Set 2'' representing the set of objects with even and odd object IDs respectively. To evaluate how well our network generalizes, we compare our results on unseen objects to the objects each network was trained on. The full breakdown of each network's scores are shown in Table~\ref{tbl:ycb_model_sets}. Although there is some performance drop on unseen objects, the gap is relatively small, showing the generalization abilities of our method. The ``Zero-Shot'' column of shows the zero-shot performance of each model on the objects it does \textit{not} see during training. 

\begin{table}[!ht]
    \small
    \centering
    \begin{tabular}{|c||c|c|c|}
    \hline
     & \multicolumn{2}{c|}{Our method trained on } & \\
    Tested on & Set 1 & Set 2 & Zero-Shot  \\
    \hline
    Object Set 1 & 0.624 & \textit{0.543} & \textit{0.543}  \\
    Object Set 2 & \textit{0.488} & 0.496 & \textit{0.488} \\
    \hline
    All Object & 0.557 &  0.520 & \textit{\textbf{0.516}} \\
    \hline
    \end{tabular}
    \caption{\label{tab:results-ycbv} AR scores on YCB-V object subsets. 
    }
    \label{tbl:ycb_model_sets}
    \vspace{-2em}
\end{table}

\subsection{Neighborhood Structure}
\label{sec:neighborhood}

We explore the effects of different neighborhood structures on the accuracy and generalization of our method. Our method uses a PointNet++~\cite{qi2017pointnet2} architecture that uses a hierarchical neighborhood structure; we compare this to a CNN architecture that uses a strict neighborhood structure  and a PointNet-based architecture~\cite{Qi2017pointnet} that uses a global structure. 
For the CNN approach, we generate a sparse difference image using the projected point differences before passing it to a ResNet18 network~\cite{he2016resnet}. Our PointNet++ approach uses normalized image coordinates for neighborhood grouping. The PointNet approach  contains the normalized image coordinates but it does not perform explicit neighborhood grouping. 
In Table \ref{tbl:score_models}, we see that the loose local neighborhood structure found in PointNet++ outperforms the global structure of PointNet as well as the strict structure used in image convolutions. 
This implies that some neighborhood structure is important for evaluating these sparse point differences, but a too strict neighborhood hampers both performance and generalization. 

\begin{table}[!ht]
\small
\centering
\begin{tabular}{|c||c|c|c|c||}
\hline
 & PointNet++ & PointNet & CNN \\
On YCB-V dataset & (Hierarchical) & (Global) & (Strict) \\
\hline
Seen (Training) Objects & \textbf{0.624}  & 0.477 & 0.533 \\
Unseen (Test) Objects & \textbf{0.488} & 0.355 & 0.386 \\
\hline
Total & \textbf{0.557} & 0.416 & 0.459 \\
\hline
\end{tabular}
\caption{Comparison of the performance of the different neighborhood structure through network architectures.}
\label{tbl:score_models}
\vspace{-2em}
\end{table}

\subsection{Input Ablations}

To determine the relative importance of each of our input channels, we retrain our networks without each dimension. We show results on YCB in Table~\ref{tbl:ycb_feature_dropout}, training on the ``Object Set 1" and testing on ``Object Set 2". Additionally, this table shows the effects of concatenating observation and model inputs (``Model without Diff"), as opposed to computing their difference (as in our method). As can be seen, using concatenation instead of differencing gives little change in performance for seen objects, whereas it gives worse performance for unseen objects. Unsurprisingly, the color information has the greatest effect on the accuracy of our system, as it is the most orthogonal to the information used by our PPF hypotheses. 

\begin{table}[ht!]
    \small
    \centering
    \begin{tabular}{|c||c|c|c|c|c|}
    \hline
    & \multicolumn{5}{c|}{Model without} \\
    & Color & Depth & Normal  & Coords & Diff \\
    \hline
    Unseen Objects & \multirow{2}*{-18\%} & \multirow{2}*{-15\%} & \multirow{2}*{-7.1\%} & \multirow{2}*{-8.9\%} & \multirow{2}*{-6.3\%} \\
    \textbf{(Zero-shot) } & & & & & \\
    \hline
    \hline
    Seen Objects & \multirow{2}*{-24\%} & \multirow{2}*{-4.2\%} & \multirow{2}*{0.8\%} & \multirow{2}*{1.1\%} & \multirow{2}*{2.1\%} \\
    (Training) & & & & & \\
    \hline
    \end{tabular}
    \caption{Percent change in AR scores on YCB Video dataset caused by removal of each input to our method. 
    }
    \label{tbl:ycb_feature_dropout}
    \vspace{-2em}
\end{table}

\subsection{Timing analysis}

\begin{figure}
    \centering
    \includegraphics[trim={15 0 25 0}, clip,width=\columnwidth]{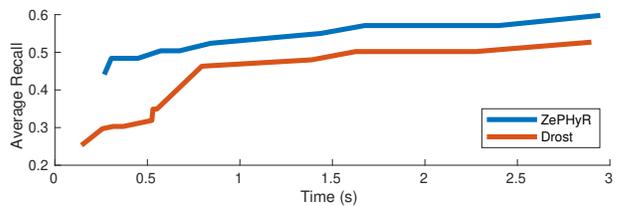}
    \caption{Speed accuracy analysis of our method (\textcolor{oceanboatblue}{blue}) over various PPF hypothesis generation hyperparameters on  LM-O. Base PPF accuracy shown in \textcolor{burntorange}{orange}. 
    }
    \label{fig:time_acc}
    \vspace{-0.5em}
\end{figure}

We analyze the inference speed of our method in Table~\ref{tbl:time_breakdown}. We separate our method into 5 stages, including generating pose hypotheses from SIFT feature matching (``SIFT"), generating pose hypotheses from PPF (``PPF"), computing, transforming and comparing the observation and model values for all hypotheses (``Projection") and inference with our scoring network (``Scoring"). Note that we only use 100 PPF hypotheses for LM-O, whereas we use additional 1000 SIFT hypotheses for YCB-V.
We found that the LM-O dataset required more accurate initial pose hypotheses, requiring significantly more processing time. To compensate for this, we evaluate the time-performance trade-off of different sets of PPF parameters on the LM-O dataset, shown in blue on Figure~\ref{fig:time_acc}. 
Since the LM-O dataset is challenging due to strong occlusions and limited scales of objects in the scene, PPF methods~\cite{drost2010ppf, vidal2018method} need a high sampling rate to produce reasonable pose estimates. Therefore, increased speed comes at the cost of performance, but our method consistently improves the accuracy of the initial hypotheses, shown in red, at all stages of the curve.

\begin{table}[ht!]
\small
\centering
\begin{tabular}{|c||c|c|c|c|c|c||}
\hline
 & SIFT & PPF & Projection & Scoring & Total \\
\hline
YCB-V & 0.142 & 0.291 & 0.051 & 0.135 & 0.619 \\
LM-O & 0 & 2.900 & 0.014 & 0.034 & 2.949 \\
\hline
\end{tabular}
\caption{Test time spent (sec) in each stage of our pipeline.}
\label{tbl:time_breakdown}
\end{table}


\subsection{Reconstructed Model Results}

To show the effectiveness of our method in robotic scenarios, we test our pipeline on newly generated object model reconstructions. Using fiducial markers~\cite{garrido2014automatic} and TSDF based surface reconstruction~\cite{zhou2013dense}, we build textured mesh model of a novel drill object. As shown in Figure~\ref{fig:Figure1}, we are able to estimate the pose of the target object while in human hands and while being manipulated by the robot; because our method is zero-shot, we do not require any retraining to estimate the pose of  new objects such as this one. We find that these poses, coupled with annotated grasp locations, allow the robot to perform task specific grasps. See our video for visualizations of the predicted poses as well as demonstration of our grasping pipeline.

\vspace{-0.25em}
\section{Conclusion}
\vspace{-0.25em}

We propose a method for zero-shot object pose estimation, focusing on pose hypothesis scoring. By extracting point differences between the projected object points and the observation and imposing a loose neighborhood structure on these points, we learn a pose scoring function that generalizes well to novel objects. On the challenging YCB-Video and LineMOD-Occlusion datasets, our method achieves state-of-the-art performance for zero-shot object pose estimation in clutter, evaluated on both textured and untextured objects. We hope that our method paves the way for roboticists to obtain accurate pose estimates for novel objects without needing additional training or data annotation.


\section{Acknowledgements}

This work was supported by NASA NSTRF, United States Air Force and DARPA under Contract No. FA8750-18-C-0092, National Science Foundation under Grant No.  IIS-1849154, and LG Electronics.







\bibliographystyle{IEEEtran}
\bibliography{IEEEabrv,root}

\clearpage

\section*{APPENDIX}

\subsection{Pose Error Metrics}

\newcommand{\Pgt}{\mathbf{P}}
\newcommand{\Pest}{\hat{\mathbf{P}}}
\newcommand{\objmodel}{\mathcal{M}}
\newcommand{\Dgt}{D}
\newcommand{\Dest}{\hat{D}}
\newcommand{\Vgt}{V}
\newcommand{\Vest}{\hat{V}}
\newcommand{\avg}{\text{avg}}
\newcommand{\norm}[1]{\left\lVert#1\right\rVert}

For the evaluation of our method, we adopt three metrics proposed by the BOP challenge 2019 \cite{hodan2018bop}.
Given an object model $\objmodel$, an estimated pose $\Pest$ and its corresponding ground truth $\Pgt$, we calculate three metrics as follows. 

\subsubsection{Visible Surface Discrepancy (VSD)}
Given the estimated and ground truth pose $\Pest$ and $\Pgt$, the object model is rendered to obtain the estimated and ground truth distance maps $\Dest$ and $\Dgt$ respectively. The distance maps are then compared with the observed distance map to obtain the visibility masks $\Vest$ and $\Vgt$, which are the sets of pixels where the object is visible in the test image. Then the VSD measures the discrepancy of the estimated and ground truth distance maps that are visible as follows. 

\begin{align}
    e_{\text{VSD}}=\underset{p\in \Vest \cup \Vgt}{\avg} 
    \begin{cases}
    0 \quad \text{if } p \in \Vest \cap \Vgt \land |\Dest(p)-\Dgt(p)| < \tau
    \\
    1 \quad \text{otherwise,}
    \end{cases}
\end{align}
where $p \in \Vest \cap \Vgt$ iterates over all pixels that is both visible under $\Pest$ and $\Pgt$. 
Note here VSD only measures geometry alignment (color agnostic) and treats indistinguishable poses as equivalent by considering only the visilbe object part. 

\subsubsection{Maximum Symmetry-
Aware Surface Distance (MSSD)}
Consider a object point cloud $\mathcal{M}=\{x_j\}$ and a set of symmetric transformations $\mathcal{T}$ for this object, MSSD is defined as
\begin{align}
    e_{\text{MSSD}} = \min_{T \in \mathcal{T}}
    \max_{x_j\in \mathcal{M}}
    \norm{\Pest x_j - \Pgt T x_j}_2.
\end{align}
MSSD measures the surface deviation in 3D, and thus is relevant for robotics applications. 

\subsubsection{Maximum Symmetry-Aware Projection Distance (MSPD)}
Let $\text{proj}$ denote the 2D projection operations. Then MSPD is defined as 
\begin{align}
    e_{\text{MSPD}} = \min_{T \in \mathcal{T}}
    \max_{x_j\in \mathcal{M}}
    \norm{\text{proj}(\Pest x_j) - \text{proj}(\Pgt T x_j)}_2.
\end{align}
Therefore MSPD measures the maximum perceivable discrepancy in 2D image space. 

Based on the above three metrics, a overall Average Recall (AR) score is computed. Given an estimated pose, it is considered correct if $e<\theta_e$ w.r.t pose error metric $e$, where $e\in \{e_{\text{VSD}}, e_{\text{MSSD}}, e_{\text{MSPD}}\}$ and $\theta_e$ is the threshold of correctness. The ratio of correctly-estimated poses over all pose estimation targets, is defined as recall. Then $\text{AR}_e$ is the Average Recall w.r.t. metric $e$, which can be calculated for multiple thresholds $\theta_e$ and multiple misalignment tolerance $\tau$ in the case of $e_{\text{VSD}}$. The final AR score is computed as the average of three:
\begin{align}
    \text{AR} = (\text{AR}_{\text{VSD}} + \text{AR}_{\text{MSSD}} + \text{AR}_{\text{MSPD}})/3. 
\end{align}

\subsection{Ground Truth Translation Results}

The Multipath Augmented Autoencoders~\cite{sundermeyer2020multipath} baseline assumes that the object is cropped from the scene prior to input.  In contrast, the focus of ZePHyR is to perform zero-shot pose estimation in cluttered scenes which contain multiple objects.  In such cluttered scenes, finding the correct object crop of a novel object is non-trivial. 

In BOP leaderboard\footnote{\href{https://bop.felk.cvut.cz/method_info/96/}{https://bop.felk.cvut.cz/method\_info/96/}}, Multipath Autoencoders~\cite{sundermeyer2020multipath} reports their performance with the assistance of a dataset-wise trained MaskRCNNs as a segmentation network. Consider that the main contribution of~\cite{sundermeyer2020multipath} is learning rotation encoding that generalizes over objects, we resolve the scale ambiguity and isolate the orientation error by providing this network with the ground truth translation for each object at test time. As shown in Table~\ref{tbl:exp_mask}, our method still outperforms~\cite{sundermeyer2020multipath}, especially on the YCB-V dataset where most objects have rotational symmetry. 

\begin{table}[h!]
\centering
\begin{tabular}{c|cc|c}
\hline
\multirow{2}{*}{Method} & \multicolumn{2}{c|}{Multipath AutoEncoder} & Ours               \\ \cline{2-4} 
                        & w/o GT trans   & w/ GT trans   & w/o GT trans \\ \hline
YCB-V                   & 0.289                & 0.355               & 0.516              \\
LM-O                    & 0.217                & 0.560               & 0.598              \\ \hline
\end{tabular}
\caption{AR scores for different method with and without ground truth translation (``GT trans''). }
\label{tbl:exp_mask}
\end{table}

\subsection{Pose Hypothesis Ablations Results}

We test our scoring method on different subsets of pose hypotheses to explore our sensitivity to the hypothesis generation method.  In Table.~\ref{tbl:exp_hypo}, we report the AR scores of the Point Pair Features baseline (``PPF'')~\cite{drost2010ppf}, our scoring method using pose hypotheses generated only from PPF (``PPF+Scoring''), our scoring method using pose hypotheses generated only from SIFT feature matching (``SIFT+Scoring'') and our scoring method using pose hypotheses generated from both PPF and SIFT (``Both+Scoring''). 
The results indicates that on the YCB-V dataset, where most objects have high-quality mesh models and rich textures, the SIFT feature matching method provides valuable pose hypotheses.  When combining PPF and SIFT hypotheses with our scoring method, the results improve over using our scoring method with PPF hypotheses alone. LineMOD (LM-O), however, contains mostly low texture or textureless objects. For this dataset, SIFT hypotheses are less useful and adding them mildly reduces the accuracy of our method but needs more processing time. 

\begin{table}[h!]
\centering
\begin{tabular}{|c|c|c|c|c|}
\hline
Method &  PPF  & PPF+ZePHyR & SIFT+ZePHyR & Both+ZePHyR \\ \hline
YCB-V  &  0.344  &   0.458     &    0.390     &     0.516        \\ \hline
LM-O   & 0.527 &    0.598    &    0.011     &     0.595     \\ \hline
\end{tabular}
\caption{BOP AR scores for ZePHyR based on different hypothesis generation methods. }
\label{tbl:exp_hypo}
\end{table}




\subsection{Network Details}

\subsubsection{PointNet++}
\label{sec:pointnetplusplue}

As mentioned in Section III-D.3, we reduce the sizes of MLP and adjust parameters of original PointNet++ design, to enable the training of the whole network with 1100 pose hypotheses in 11 GB GPU memory. We uniformly downsample the object mesh models so that the leaf size for the voxel grid is 7 millimeter and each object has 1000 points on average, and further randomly subsampled the input points down to $2000$ when the number of points in the downsampled object model still exceeds this number. The detailed network architecture is described as follows.  

We use the single scale grouping (SSG) version of PointNet++. 
Following architecture protocol in~\cite{qi2017pointnet2}, we denote SA($K$,$r$,[$l_1,...,l_d$]) as a set abstraction (SA) level with $K$ local regions of ball radius $r$ using PointNet of $d$ fully connected layers with width $l_i$ ($i=1,...,d$). SA([$l_1,...l_d$]) represents a global set abstraction level that converts set to a single vector. FC($l$,$dp$) represents a fully connected layer with width $l$ and dropout ratio $dp$. All fully connected layers are followed by batch normalization~\cite{ioffe2015batch} and ReLU activation functions, except for the last score prediction layer. The resulting PointNet++ architecture is as follows: 
\begin{equation*}
\begin{split}
    & SA(128,0.2,[16, 32]) \rightarrow SA(16,0.5,[32. 64]) \rightarrow \\ & SA([64, 128]) \rightarrow  FC(64,0.4) \rightarrow FC(16,0.4) \rightarrow FC(1)
    \end{split}
\end{equation*}

\subsubsection{PointNet}

For the ablation experiment on PointNet in Section V-C, we also use a reduced version of Classification Network described in~\cite{Qi2017pointnet}. We remove the input transform and feature transform layers. We use a three-layer MLP, with the size of the hidden layer to be $16$, pre-bottleneck, a bottleneck max pooling layer of dimension $16$, and a 3-layer MLP with the hidden layer size $64$ post-bottleneck. All except the last MLP layers are followed by a batch normalization layer~\cite{ioffe2015batch} and a ReLU activation. The final output of the last layer estimates a single score for each input point cloud. 

\subsubsection{Convolutional Network}

For the CNN mentioned in Section V-C, we use a vanilla ResNet-18~\cite{he2016resnet} with no pretrained-weight. The the number of input channels of the first layer is expanded to match the number of error features, and the last layer is changed to a 2-layer MLP with the hidden layer size 64. The final output is a single score for each pose hypothesis. 

\subsection{Training Details}

For computational efficiency, we subsample the training data points in the YCB-V and LM-O datasets and pre-process them for fast training. Specifically, from the YCB-V training split, we evenly sampled $4716$ observations, containing $2346$ observations of objects with even IDs and 2370 of objects with odd IDs. 
From the synthetic training set of LineMOD dataset~\cite{hodan2019photorealistic}, we evenly sampled 1749 observations of objects that are not in LM-O dataset as the training set. 
The observations of the training objects are then split, with $90\%$ used for training and 10\% used for validation. 
After training, the model weights at the epoch with lowest error on validation set of the ``seen'' objects are selected for evaluation, and the observations of ``unseen'' objects are not used during training or validation. 

To train the PointNet and PointNet++ archetectures, we use an Adam optimizer~\cite{kingma2014adam} with an initial learning rate $3\times 10^{-4}$. For the CNN training, the initial learning rate is $1\times 10^{-5}$. We trained each network for 100 epochs and the learning rate reduces to $1/10$ after epoch 30 and 80. 

We augment the training data by randomly jittering the brightness, contrast, saturation and hue of the observation images by factor of $0.2$, $0.2$, $0.2$ and $0.05$ respectively. To prevent overfitting to the training objects, we also jointly perturb the color of the model and the observation color, changing the color of both the real and rendered data in the same way. The factors for brightness, contrast, saturation and hue in this process are all $0.5$. 

\subsection{Comparison of DeepIM}

\begin{table}[h!]
    \centering
    \begin{tabular}{|c|c|c|c|}
    \hline
        Method & Drost~\cite{drost2010ppf} & Drost~\cite{drost2010ppf} + & Drost~\cite{drost2010ppf} + \\
         &  & DeepIM~\cite{li2018deepim} & ZePHyR(ours) \\
        \hline
        YCB-V & 0.344 & 0.324 & 0.516 \\
        LM-O & 0.527 & 0.165 & 0.598 \\
    \hline
    \end{tabular}
    \caption{BOP AR scores for DeepIM taking pose hypothesis from PPF.}
    \label{tab:deepim}
\end{table}

ZePHyR is a pose hypothesis scoring method, which is different from other learned pose refinement methods in the literature in the sense that ZePHyR can select the best one among multiple pose hypotheses over the entire search space while pose refinement only improves a single pose in a local region. To quantify this difference, we evaluate DeepIM~\cite{li2018deepim} using the publicly available implementation\footnote{\hyperlink{https://github.com/NVlabs/DeepIM-PyTorch}{https://github.com/NVlabs/DeepIM-PyTorch}} and the model checkpoint trained on the YCB-V dataset (model trained on LM-O is not provided). 
Note that for YCB-V, the DeepIM performance is not zero-shot as YCB objects are seen during training, while its performance on LM-O is zero-shot. 
The pose for DeepIM is initialized from the results of Drost~\etal~\cite{drost2010ppf}, and the results are shown in Table.~\ref{tab:deepim}. 

According to Drost~\etal~\cite{drost2010ppf}, ICP algorithm~\cite{besl1992icp} is already used as a post-processing step in the PPF pipeline and thus their pose estimation results are already very accurate geometrically. Therefore, the room of improvement left for DeepIM is very limited as it only refines on a single pose hypothesis in the local region. 
We observe that for seen objects in the YCB-V dataset, DeepIM does not make obvious improvement based on Drost and even makes it slightly worse. And when DeepIM is tested on unseen objects (trained on YCB objects and tested on the LM-O dataset), it makes the initial estimation drastically worse. 

\subsection{Qualitative Results}

\begin{figure*}[p]
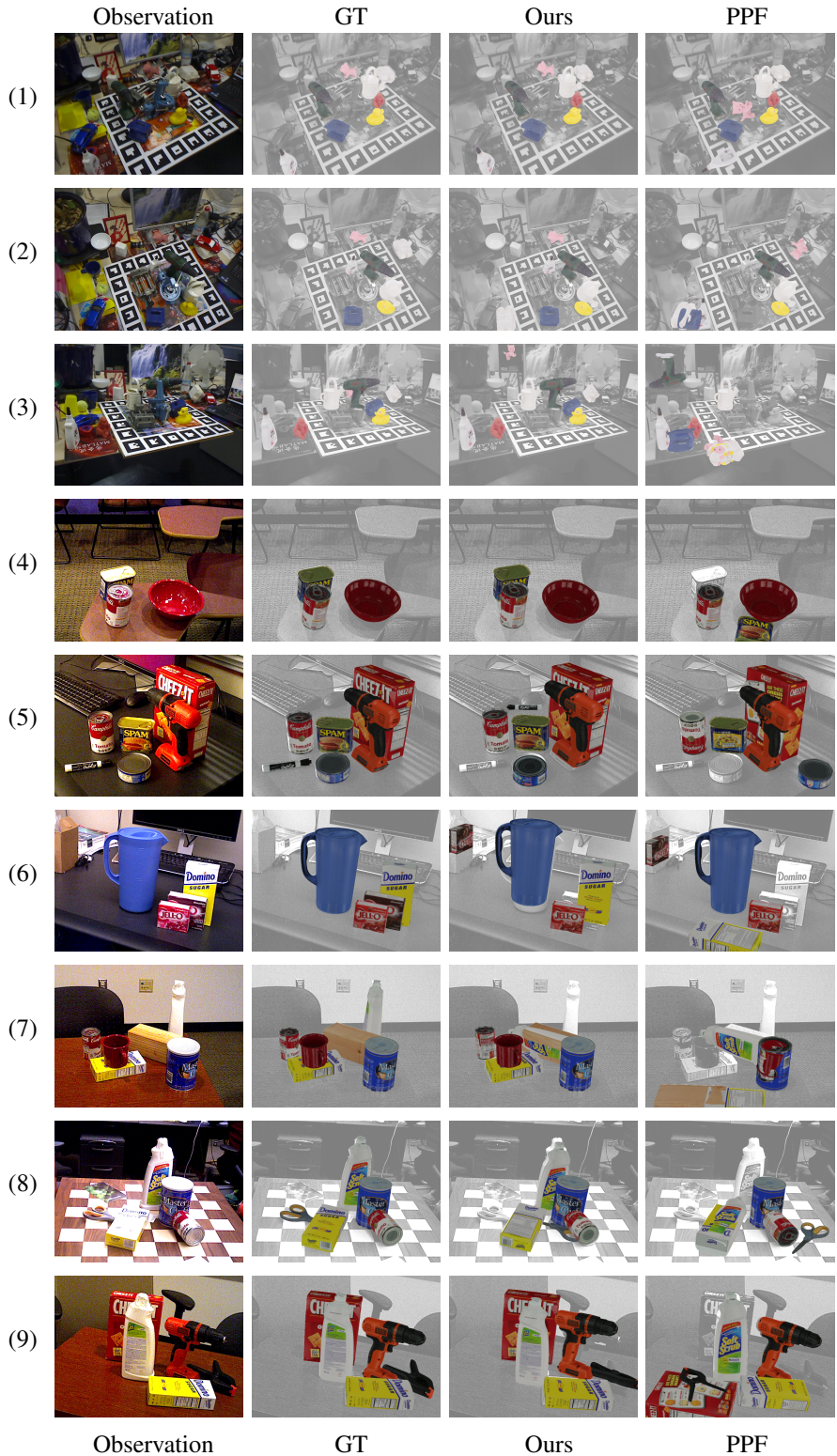

\def \qualList {
{lmo/000002_000446},
{lmo/000002_000039},
{lmo/000002_000791},
{ycbv/000053_000138},
{ycbv/000059_001068},
{ycbv/000058_000670},
{ycbv/000055_000360},
{ycbv/000051_000649},
{ycbv/000054_001042}%
}

\def \qualListFail{
{ycbv_010_000050_000620},
{lmo_009_000002_000266}%
}
\centering
\hspace{0.025\textwidth}
\begin{tabularx}{0.63\textwidth}{YYYY}
Observation & GT & Ours & PPF 
\end{tabularx}
\\
\foreach \qualName [count=\x from 1] in \qualList {
\centering
(\x)
\raisebox{-.5\height}{
\includegraphics[width=0.15\textwidth]{figures/qual/\qualName_img.png}   \includegraphics[width=0.15\textwidth]{figures/qual/\qualName_gt.png} 
\includegraphics[width=0.15\textwidth]{figures/qual/\qualName_score.png} 
\includegraphics[width=0.15\textwidth]{figures/qual/\qualName_ppf.png} 
}
\vspace{0.2cm}
\\
}
\centering
\hspace{0.025\textwidth}
\begin{tabularx}{0.63\textwidth}{YYYY}
Observation & GT & Ours & PPF \\
\end{tabularx}
\caption{Qualitative results on LM-O (first 3 rows) and YCB-V (last 6 rows) dataset. Raw input image and ground truth renders shown in the first and second column, respectively. The third and fourth column compare the top results using our scoring pipeline (``Ours'') and the original PPF (``PPF'') hypothesis algorithm~\cite{drost2010ppf}, respectively. }
\label{fig:qual_results_appx}
\end{figure*}

Figure~\ref{fig:qual_results_appx} shows the qualitative results of both our method and the baseline over the YCB-V and LM-O datasets. The left column shows the full scene; the second column shows the ground-truth pose for the target object.  The third column shows the highest-scoring pose according to our method, and the last column shows the highest-scoring pose according to the PPF baseline~\cite{drost2010ppf}.  In the 3rd and 4th columns, the selected pose hypothesis for each method is rendered into the frame. 

Overall, Our method demonstrates a better performance than the PPF baseline. As PPF only considers geometry, it cannot determine the correct orientation on some objects that are symmetrical in shape but have distinguishing texture, like the ``Master Chef'' can and tomato can in row (5), (7) and (8) in Figure~\ref{fig:qual_results_appx}. But our method considers both shape and color information, and thus can make correct estimations in such cases. PPF also tends to match the flat side of an object to the flat top of a table, such shown in row (3), (6), (7) and (9) in Figure~\ref{fig:qual_results_appx}; our method fixes such errors. 

Figure~\ref{fig:qual_results_appx} also shows some cases where our method fails. In row (8), due to the over exposure on the surface of the sugar box, our method mixes the back side of the box with the front side. In row (7), our method fails to detect the ``Soft Scrub'' bottle probably because only its side is facing towards the camera, where almost no texture or color information is present. The toy cat in row (3) and the egg box in row (2) are two failure cases where the occlusion is so strong that the whole object is almost invisible. 

\subsection{Failure Case Analysis}

\begin{figure*}[p]

\includegraphics[width=0.32\textwidth]{figures/failure/img.png}
\includegraphics[width=0.32\textwidth]{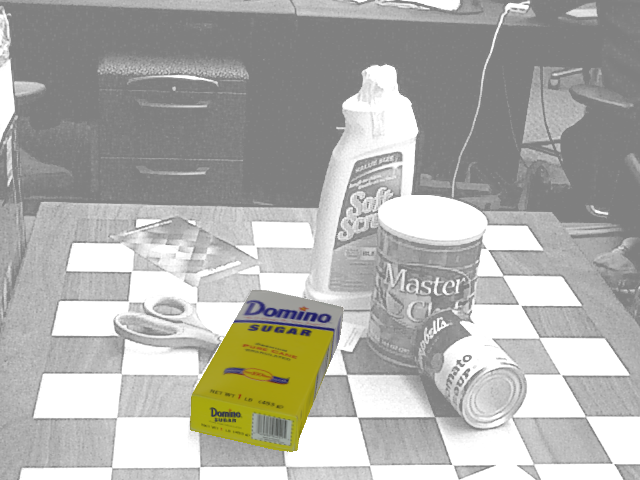} 
\includegraphics[width=0.32\textwidth]{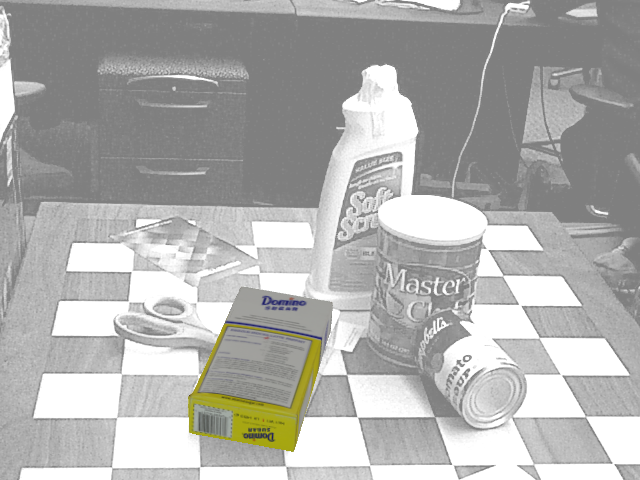} 
\begin{tabularx}{\textwidth}{YYY}
(a) Observation & (b) Best & (c) Ours \\
\end{tabularx}

\includegraphics[width=\textwidth]{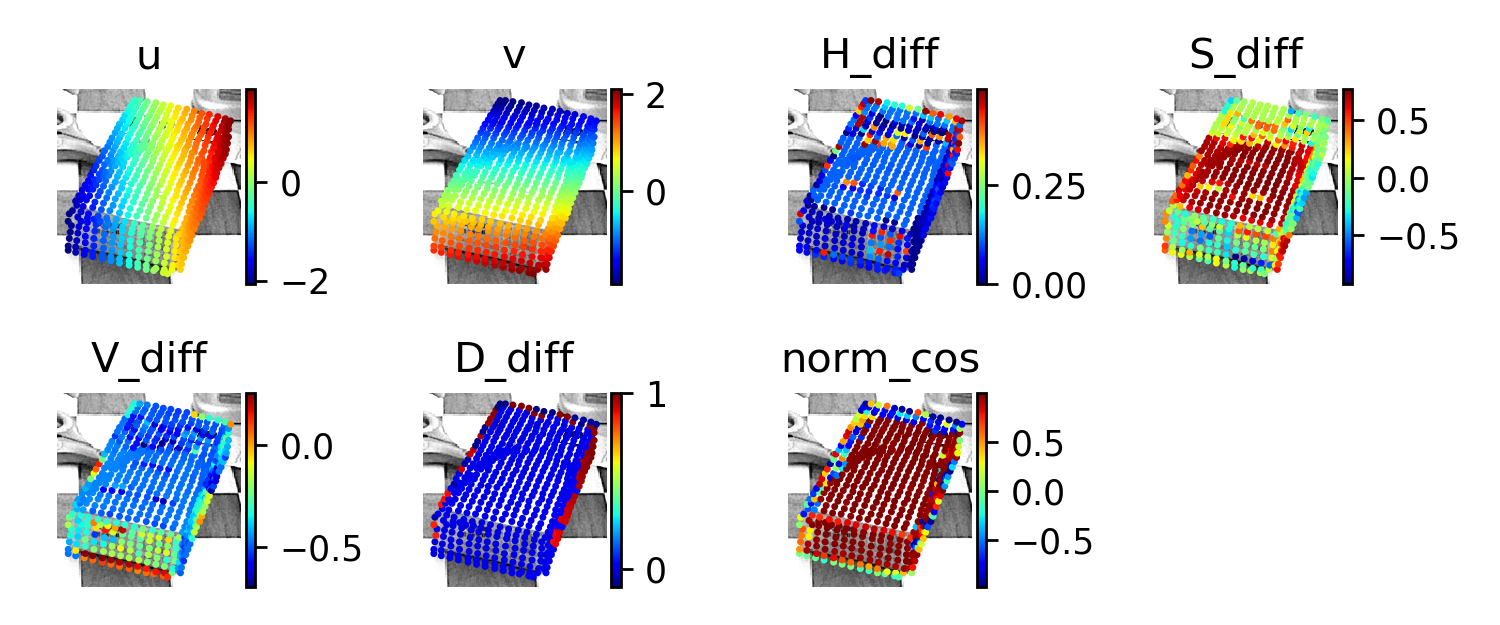}
\begin{center}
    (d) Error features for the best hypothesis
\end{center} 

\includegraphics[width=\textwidth]{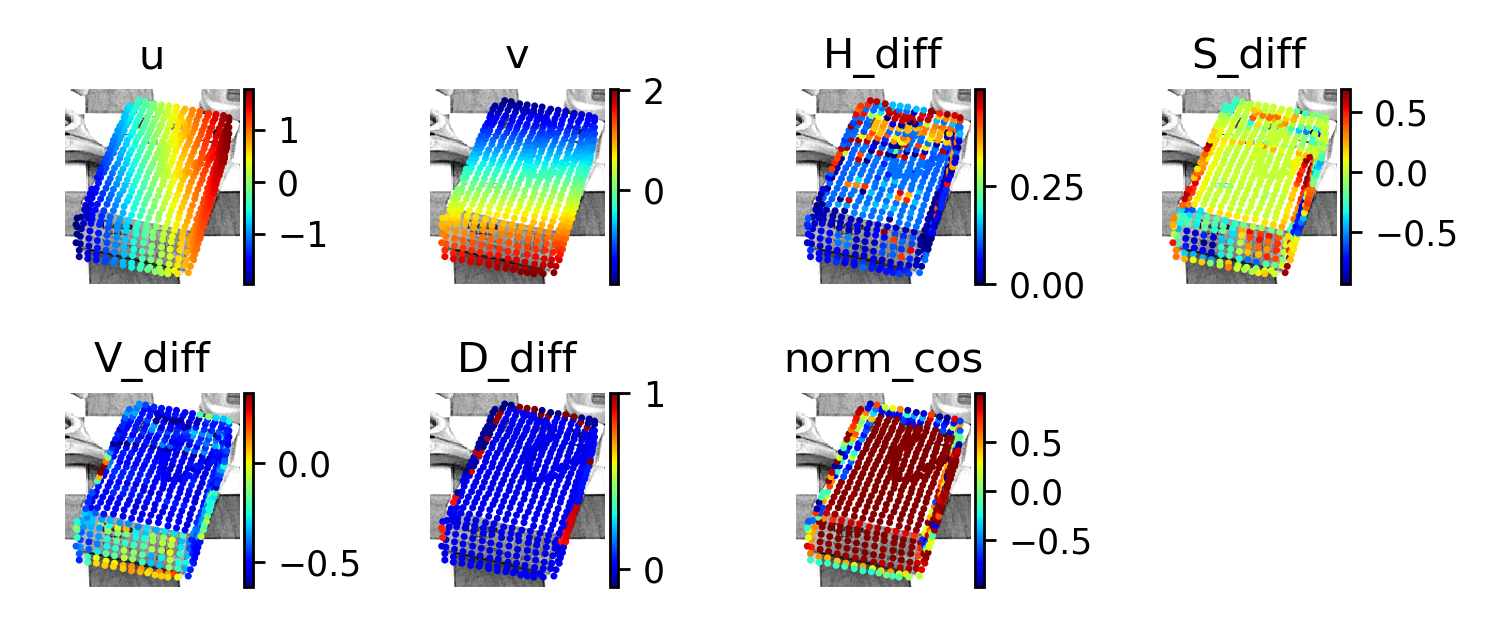}
\begin{center}
    (e) Error features for our result
\end{center} 

\caption{Failures case of our method. ``Best'' means the pose that has the lowest ADD error in the pose hypothesis set. ``Ours'' means the highest scoring hypothesis returned by our method. In plot (d) and (e), ``u'' and ``v'' are  the normalized projection coordinates. ``H\_diff'', ``S\_diff'', ``V\_diff'' and ``D\_diff'' represent the signed difference of the hue, value, saturation and depth between projected model points and the observation respectively. ``norm\_cos'' is the cosine of the angle between transformed model normal vectors and observed normal vectors. }
\label{fig:failure_features}
\end{figure*}

Figure~\ref{fig:failure_features} further elaborates the failure case of the sugar box in the row (8) of Figure~\ref{fig:qual_results_appx}. As we can see, due to the reflection, the upper surface of the sugar box in the observation is overly lightened, which makes the saturation and value errors of the wrongly-picked hypothesis smaller than those of the correct one. However, our method correct recover the geometry and still presents a reasonable result. 

\subsection{Time-Accuracy Trade-off on LM-O dataset}

In Table~\ref{tbl:ppf-detail}, we report the detailed data for the time-accuracy trade-off curve in Figure 4 in the main paper. We here only vary the PPF parameters and thus its inference time. The speed of our scoring network (ZePHyR) is unchanged. In the table, ``Model SD'' and ``Scene SD'' are the sampling distance on the model point cloud and the scene point cloud respectively, relative to the model diameter. Higher numbers lead to smaller point clouds and faster processing times. ``Ref Pt Rate'' is the ratio of the points on the scene point cloud that are used as reference points when sampling point pairs~\cite{drost2010ppf}. ``Dense Object PC'' means the input object model to PPF is directly converted from the mesh model without downsampling. ``Sparse Object PC'' means PPF uses the downsampled object point cloud that is used in the scoring network, as described in Section~\ref{sec:pointnetplusplue}. ``Sparse'' and ``Dense'' in ``Refinement'' column indicates the spacial density of the point cloud used for ICP step in PPF. We refer readers to \cite{drost2010ppf} and \cite{halcon2019surface} for more details. 

Note that ZePHyR is a scoring network on the provided pose hypotheses, and in the table, our PPF+ZePHyR demonstrate a constant improvement over the PPF baseline by a large margin with only little time overhead. This means our method is able robustly pick better hypothesis from the PPF's output. Comparing the first and the third row in the table, we can find that PPF+ZePHyR achieves comparable results with PPF but is sped up by more than 3 times. 


\begin{table*}[p]
\centering
\begin{tabular}{ccccc|cccc}
\hline
Model SD & Scene SD & Ref Pt Rate & Object PC & Refinement & Time (PPF) & BOP score (PPF) & \begin{tabular}[c]{@{}c@{}}Time \\ (PPF+ZePHyR)\end{tabular} & \begin{tabular}[c]{@{}c@{}}BOP score \\ (PPF+ZePHyR)\end{tabular} \\ \hline
0.03     & 0.03     & 1           & Dense     & Dense      & 2.900      & 0.527           & 2.948                                                        & \textbf{0.598}                                                    \\
0.03     & 0.05     & 1           & Dense     & Sparse     & 1.626      & 0.502           & 1.674                                                        & 0.571                                                             \\
0.05     & 0.05     & 1           & Dense     & Sparse     & 1.388      & 0.480           & 1.436                                                        & 0.550                                                             \\
0.05     & 0.05     & 0.5         & Dense     & Sparse     & 0.794      & 0.463           & 0.842                                                        & 0.524                                                             \\
0.05     & 0.07     & 0.5         & Dense     & Sparse     & 0.530      & 0.349           & 0.578                                                        & 0.456                                                             \\
0.03     & 0.04     & 0.5         & Sparse    & Sparse     & 0.524      & 0.319           & 0.572                                                        & 0.504                                                             \\
0.05     & 0.07     & 0.25        & Dense     & Sparse     & 0.315      & 0.303           & 0.363                                                        & 0.408                                                             \\
0.03     & 0.04     & 0.2         & Sparse    & Sparse     & 0.257      & 0.297           & 0.305                                                        & 0.484                                                             \\
0.03     & 0.05     & 0.2         & Sparse    & Sparse     & 0.219      & 0.253           & 0.267                                                        & 0.441                                                             \\
0.05     & 0.05     & 0.2         & Sparse    & Sparse     & 0.200      & 0.213           & 0.248                                                        & 0.379                                                             \\ \hline
\end{tabular}
\caption{Inference time and performance on the LM-O dataset of PPF and PPF+ZePHyR using different PPF settings. }
\label{tbl:ppf-detail}
\end{table*}

\end{document}